\documentclass[runningheads]{llncs}
\usepackage[T1]{fontenc}
\usepackage{multirow}
\usepackage{amsmath}
\usepackage{graphicx}
\begin{document}
\title{Self-supervised Auxiliary Learning for Texture and Model-based Hybrid Robust and Fair Featuring in Face Analysis}
%
%
\author{Shukesh Reddy$^1$  \and
Nishit Poddar$^1$ \and Srijan Das$^2$ \and
Abhijit Das$^1$}
\authorrunning{S. Reddy et al.}
\institute{ $^1$ Machine Intelligence Group, Department of CSIS,\\ Birla Institute of Technology and Science, Pilani, Hyderabad Campus, India  
\\
$^2$ University of North Carolina at Charlotte, United States
\email{abhijit.das@hyderabad.bits-pilani.ac.in}}

\maketitle            

\begin{abstract}
In this work, we explore Self-supervised Learning (SSL) as an auxiliary task to blend the texture-based local descriptors into feature modelling for efficient face analysis. Combining a primary task and a self-supervised auxiliary task is beneficial for robust representation. Therefore, we used the SSL task of mask auto-encoder (MAE) as an auxiliary task to reconstruct texture features such as local patterns along with the primary task for robust and unbiased face analysis. We experimented with our hypothesis on three major paradigms of face analysis: face attribute and face-based emotion analysis, and deepfake detection. Our experiment results exhibit that better feature representation can be gleaned from our proposed model for fair and bias-less face analysis. 

\end{abstract}
\section{Introduction}
\vspace{-4mm}
Achieving accuracy in face analysis is heavily dependent on using efficient face description methods. The research of facial representation techniques has been comprehensive, encompassing principal component analysis (PCA)\cite{139758}, linear discriminant analysis (LDA)\cite{Etemad:97}, local binary patterns (LBP)\cite{10.1007/978-3-540-24670-1_36}, and local directional patterns (LDP)\cite{5597095}. Such holistic approaches like PCA, LDA, focus on deriving global information. On the other hand, local descriptor techniques like LBP and LDP have gained popularity because of their superior capability to accurately handle certain local features. Such techniques may face challenges when dealing with complex changes such as facial expressions, posture, and illumination. 

In the past few years, Deep learning models such as AlexNet\cite{NIPS2012_c399862d}, GoogLeNet\cite{szegedy2014goingdeeperconvolutions}, and ResNet\cite{7780459} have revolutionized the field of facial recognition. They have produced limited results in major competitions such as the ILSVRC. However, the application of these models in real-world, data-rich environments is difficult to achieve. This is sometimes hampered by the need for large amounts of training data and fast processing resources. FaceNet\cite{Schroff_2015} has solved this problem by implementing an optimal algorithm that uses deep variable networks to generate facial input with a consistent learning curve. Local features for CNN have been explored in a few works \cite{juefei2017local} and also for face analysis \cite{ravi2020face}. 
Further in the recent past, the Vision Transformer (ViT) architecture-based featuring has been explored for faces\cite{das-limiteddatavit-wacv2024}. This work also utilizes SSL as an auxiliary task to learn from limited. However, a more robust path for bleeding local and model features for face analysis is not explored in the literature. This motivates us to propose a hybrid approach that combines the advantages of texture-based and model-based features with SSL for robust and fair face analysis.\\

\noindent The specific contribution of this work is as follows:

\begin{enumerate}
    \item We proposed a Self-supervised Learning (SSL) based auxiliary task to blend the texture-based local descriptors into feature modelling for efficient face analysis.
    \item We made a deep analysis of the different ways to combine the texture-based local descriptors into feature modelling for efficient and unbiased face analysis.
    \item A detailed benchmark experiments on different types of task-on-face analysis with different ways to combine the texture-based local descriptors into feature modelling.
\end{enumerate}

\section{Related Work}
\vspace{-3mm}
\textbf{Vision Transformers}-Recent advances in Vision Transformers (ViTs) have led to the development of numerous models for various applications \cite{arnab2021vivitvideovisiontransformer,bertasius2021spacetimeattentionneedvideo,cheng2021perpixelclassificationneedsemantic,dosovitskiy2021imageworth16x16words,liu2021videoswintransformer,ranasinghe2022selfsupervisedvideotransformer,NEURIPS2021_6a30e32e,Wang_2022,xie2021segformersimpleefficientdesign,zhou2021deepvitdeepervisiontransformer,zhu2021deformabledetrdeformabletransformers}. For best performance, these models need extensive datasets and pre-training. The DeiT model was enhanced to decrease these requirements by using sophisticated regularization, data augmentation, and token extraction from convolutional layers \cite{touvron2021trainingdataefficientimagetransformers}. The tokenization technique employed by T2T \cite{yuan2021tokenstotokenvittrainingvision} was used to identify and record local structural information. Convolutional filters were used in other models to include inductive bias \cite{dai2022mstctmultiscaletemporalconvtransformer}. Hierarchical transformers  \cite{fan2021multiscalevisiontransformers,li2022mvitv2improvedmultiscalevision,9710580,wang2021pyramidvisiontransformerversatile} have integrated inductive bias by decreasing token quantities via patch merging. However, they still need datasets of moderate size for pre-training \cite{park2022visiontransformerswork}.
\\
\\
\textbf{Self-supervised Learning-}
The objective of self-supervised learning (SSL) is to create visual representations without the need for labelled data. SimCLR\cite{pmlr-v119-chen20j} and MoCo \cite{he2020momentumcontrastunsupervisedvisual} are two examples of contrastive algorithms that aim to decrease the distance between enhanced versions of the same picture (positive pairs) and maximize the distance between different images (negative pairs). Optimal non-contrastive techniques, such as BYOL \cite{byol}and DINO \cite{dino}, just aim to decrease the distance between positive pairings. Models based on reconstruction \cite{mae,convmae,simmim,videomae} employ an encoder to produce latent representations and a decoder to reconstruct the picture. Supervised learning (SSL) is extensively employed for the extensive pre-training of Vision Transformers (ViTs), therefore improving their performance on subsequent assignments.
\\ \\
\textbf{ViTs for small datasets-}Liu et al.~\cite{efficient}  introduced an additional self-supervised task to enhance the resilience of vision-to-text (ViT) training on smaller datasets. This job entails the prediction of relative distances between tokens and is trained in conjunction with primary tasks. In contrast, Li et al. ~\cite{local}  performed distillation in the hidden layers of ViT extracted from a lightweight CNN-trained model. Lee et al. ~\cite{small_data_wacv} came up with a ViT architecture that uses shifting patch tokenization and locality self-attention to make up for the lack of location inductive bias. The SSL+Fine-tuning approach, presented by Gani et al.~\cite{smalldata}, shares similarities with the pretext task in DINO ~\cite{dino}. These techniques obviate the necessity for extensive pre-training and enable ViTs to acquire effective representations with a restricted amount of input.
\\ 

\textbf{Recent Works on Facial Analysis-}Facial analysis encompasses several tasks such as face parsing\cite{7780765,zheng2022decoupledmultitasklearningcyclical}, landmark detection \cite{lan2022hihaccuratefacealignment,zhou2023starlossreducingsemantic}, head posture estimation\cite{Zhang_2023_CVPR}, facial attribute recognition\cite{miyato,noroozi2017unsupervisedlearningvisualrepresentations,Shu_2021_CVPR}, age or gender or race and bias related to the same prediction\cite{Cao_2020,kuprashevich2023mivolomultiinputtransformerage,7301352,li2021learningprobabilisticordinalembeddings,das2018mitigating} and landmark visibility\cite{Kumar_2020_CVPR,9523977}. Tasks such as face swapping\cite{cui2023facetransformerhighfidelity}, editing\cite{Zhu_2020_CVPR}, occlusion removal\cite{10.1109/FG57933.2023.10042570}, 3D face reconstruction\cite{wood20223dfacereconstructiondense}, driver assistance systems\cite{Head_Pose_Driver}, human-robot interaction\cite{human_robot}, retail analytics\cite{real_time_gender_age}, face verification\cite{NIPS2014_e5e63da7,Taigman_2014_CVPR}, and image generation\cite{yan2016attribute2imageconditionalimagegeneration} are crucial for many applications. Specialised models flourish in certain tasks but encounter difficulties in generalising because of task-specific pre-processing \cite{lin2019faceparsingroitanhwarping}. Several methodologies \cite{7952703,ming2019dynamicmultitasklearningface,10.1007/978-3-319-10599-4_7,zhao2021deepmultitasklearningfacial} multitask by utilizing supplementary tasks to augment the primary activity and behaviour characterization also exists \cite{happy2019characterizing,das2021bvpnet,happy2020apathy,das2021spatio,niu2019robust}. Approaches such as HyperFace \cite{8170321} and AllinOne \cite{7961718}use complementary tasks such as landmark detection and head posture estimation, while still depending on techniques like selective search in R-CNN for face identification \cite{6909475,6126456}. Several works on attack detection and generation on faces can be also found in the literature\cite{das2021demystifying,av2024latent,rachalwar2023depth,kuckreja2024indiface,roy20223d,balaji2023attending}. 
\\ 
\vspace{-5mm}
\section{Proposed Methodology}
\vspace{-4mm}
Transformer models, like ViTs, lack assumptions and prior beliefs that help discover patterns from limited data, making training difficult. Previous research on self-supervised Auxiliary Task (SSAT)\cite{das-limiteddatavit-wacv2024} suggests that optimizing ViTs for the primary job and an auxiliary work simultaneously might provide unexpectedly good results, especially when training data is low. 

\subsection{Preliminaries}
\vspace{-3mm}
We proceed to enlist the backbone and the different features used in this subsection. 

\noindent \textbf{SSAT:} {Overview of self-supervised Auxiliary Task (SSAT) \cite{das-limiteddatavit-wacv2024} }
A novel joint optimization framework is introduced, which combines the basic classification problem of ViT with a self-supervised task (SSAT). This framework enables ViT to effectively capture inductive biases from the data without the need for separate labels (see Fig 1). 

The input X, which can be either a video or an image, is subjected to data augmentations A(X) and Ã(X), in the ViT. A(X) represents a complete image or video to be used for the primary task of classification. Ã(X) is subjected to masked operation using the Masked Auto encoder (MAE) approach for images, and video masked auto encoders (VideoMAE) approach for videos. An image classification job will utilize the whole video or image, whereas an image or video reconstruction work will use the masked video or picture. We rearrange the output g(f(Ã(X)) to create the rebuilt picture and then calculate the normalized Mean Square Error (MSE) loss (SSAT) between the original and reconstructed image. The classifier process generates an output g(f(A(X)) that is used to calculate the classification loss, as measured by cross-entropy. The ViT training entails the simultaneous optimization of losses from each of these objectives. An outline of SSAT is illustrated in Figure 1.

\begin{figure*}[t]
  \center
  \includegraphics[width=10cm]{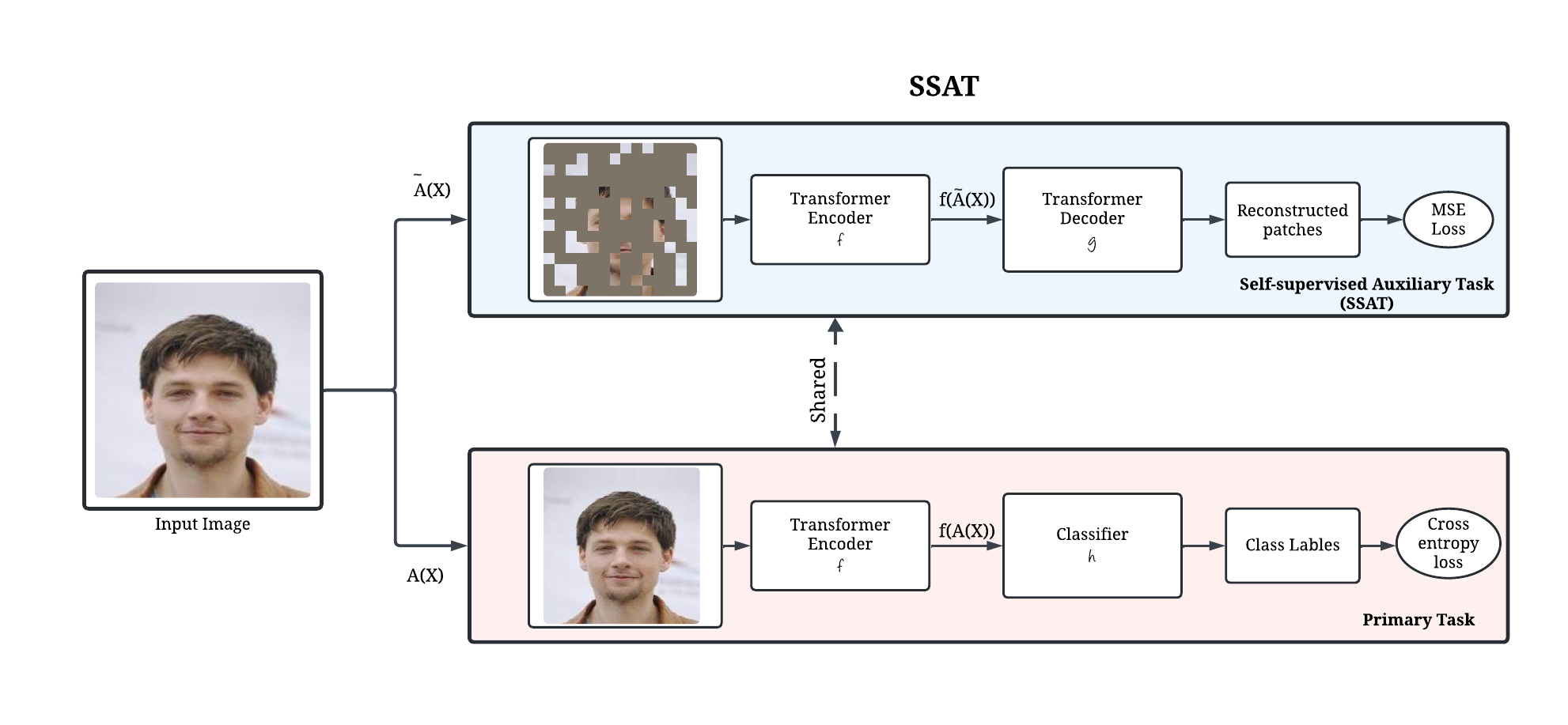}
  \vspace{-5mm}
  \caption{Self-supervised Auxiliary Task (SSAT)}
\end{figure*}

\noindent \textbf{LDP:} 
The Local Directional Pattern (LDP) \cite{LDP} describes local characteristics in a picture. Each pixel point's edge response values in all eight directions are utilized to construct an LDP feature and a relative strength magnitude code. Using a local neighbourhood to determine each byte of code sequence provides resilience in noisy conditions. To describe the image, the LDP feature is aggregated over the input picture to create the image descriptor. Face recognition using this durable LDP feature descriptor is effective even with non-monotonic light variations and random noise. Log-likelihood, Chi-square, and weighted ($X^2_w$) statistics are used to emphasize the importance of eye, nose, and mouth regions. The classifier can generalise properties better using this method. 
\begin{equation}
\chi_{w}^{2} = \sum_{i,\tau} w_{i} \frac{(S_{i}(\tau) - M_{i}(\tau))^{2}}{S_{i}(\tau) + M_{i}(\tau)}
\end{equation}
\noindent \textbf{LBP:} Wang and He \cite{ojala1994performance} introduced a model of texture analysis based on the characteristic of the texture spectrum known as the Linear binary pattern (LBP). It is represented by eight elements, each of which has one of three possible values (0,1,2) calculated from a neighbourhood of 3x3 pixels. The occurrence of the distribution of texture units computed over a region is known as the texture spectrum.
\vspace{-3mm}
\subsection{Proposed Architecture}
\vspace{-2mm}
A novel approach is introduced, which leverages the local pattern feature and facial analysis from inputted images or videos, which significantly helps to improve the performance of the network.

\noindent \textbf{Local pattern-SSAT:} The proposed framework is divided into two main tasks: upstream and downstream tasks. These tasks are used to train the main classification job of ViT-B, together with a self-supervised auxiliary task (SSAT). RGB image B(X) is used as input for the downstream job, where the Transformer encoder f and Classifier h analyze the whole image patches A(X) to calculate the classification loss \( L_{\text{downstream}} \)

A Local pattern feature A(X) is used as input for an upstream job, which involves random masking of patches in the input image X using an operation M = Ã(X). A Video video-masked auto-encoder (VideoMAE) \cite{videomae} is used to capture and rebuild the missing pixels in the sample M. The unmasked tokens are processed by Transformer encoder f to generate a latent representation f(Ã(X)) for these tokens. In conjunction with the classifier h, SSAT employs a shallow decoder g to reconstruct the image pixels that are not yet visible from the hidden representation of the observed tokens f (Ã(X)).To the decoder, a learnable masked token and the latent representation of the visible tokens f (Ã(X)) are provided. The output of the decoder maps each token representation in a linear manner onto a vector of pixel values that constitute a patch. The modelled output g(f (Ã(X))) is used to generate the reconstructed image. The loss statistic \( L_{\text{upstream}} \) is computed to determine the normalized Mean Square Error (MSE) between the original and reconstructed picture.

The whole framework serves two main tasks: an auxiliary job, which is the Local pattern feature reconstruction of inputted picture A(X) to extract local features, and a primary goal, which is the classification of inputted Original RGB image B(X). The present architecture is optimized by combining the losses incurred by the primary job and auxiliary task. Hence, the overall loss is calculated by

\begin{equation}
   L = \lambda \cdot L_{\text{downstream}} + (1 - \lambda) \cdot L_{\text{upstream}} 
\end{equation}

Where \(\lambda\) is the loss scaling factor, \(L_{\text{downstream}}\) is the MSE loss, \(L_{\text{upstream}}\) is the cross-entropy loss.An overview of our framework is depicted in Figure 2

We adopted the flowing combination of local pattern feature and RGB as input for our model as shown in Figure 2:

\begin{itemize}
    \item Local pattern feature as input to reconstruct RGB and RGB to classifier (LRRnRC)
    \item Local pattern feature as input to reconstruct RGB and Local pattern feature to classifier (LRRnLC)
    \item RGB as input to reconstruct Local pattern feature and RGB to classifier (RRLnRC)
    \item RGB as input to reconstruct Local pattern feature, RGB and RGB to classifier (RRLRnRC)
    \item RGB as input to reconstruct RGB and RGB to classifier (RRRnRC)
\end{itemize}

\vspace{-5mm}

\begin{figure*}[t]
  \centering
  \includegraphics[width=12.5cm]{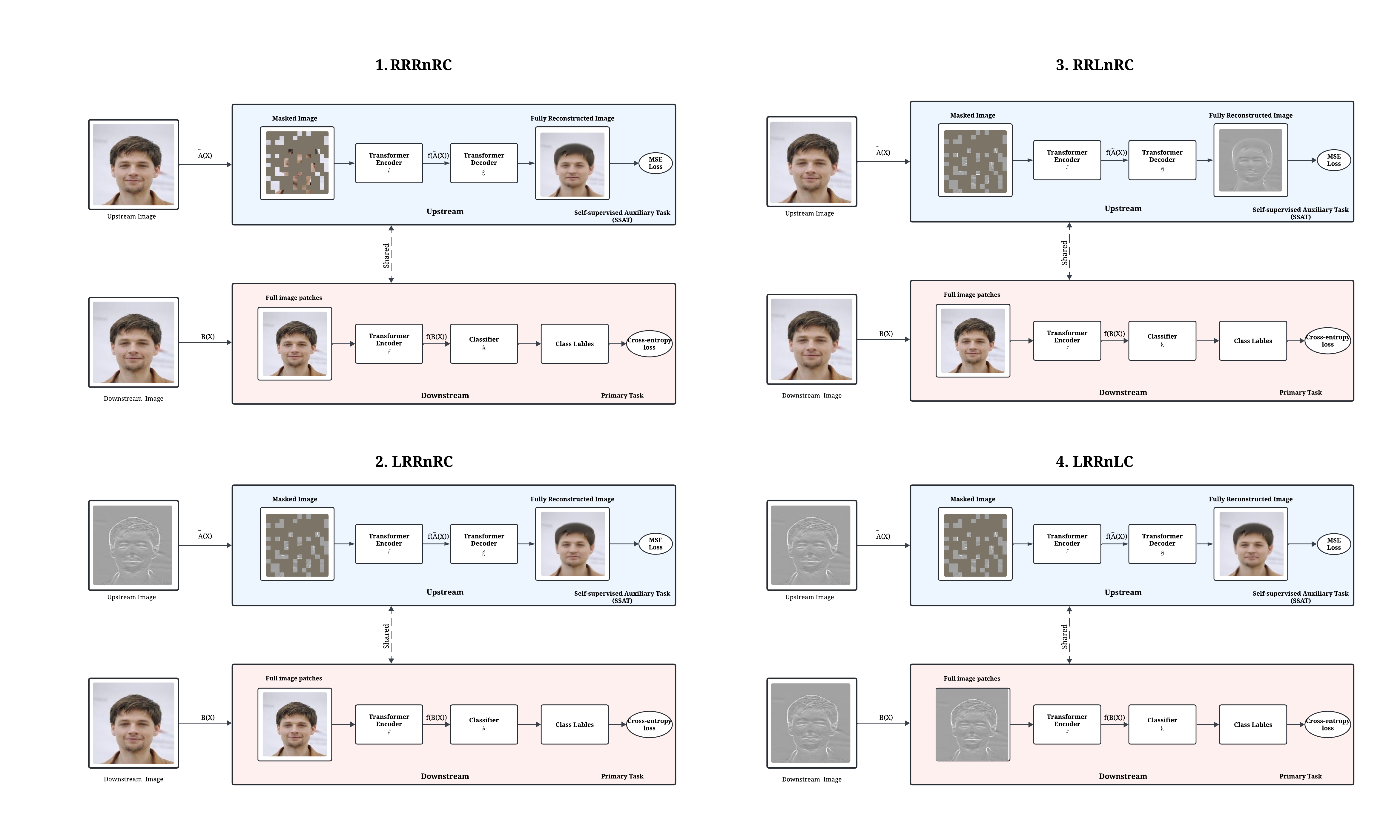}
  \vspace{-4mm}
  \caption{LDP-Self-supervised Auxiliary Task (SSAT) Experiments}
\end{figure*}
\section{Experimental Analysis}
\vspace{-3mm}
A description of the datasets utilized is provided in Section 4.1. Section 4.2 includes documentation of the implementation details and hardware utilized. Section 4.3 describes the experimental results and analysis. 
\begin{figure*}[h!]
  \centering
  \includegraphics[width=12cm]{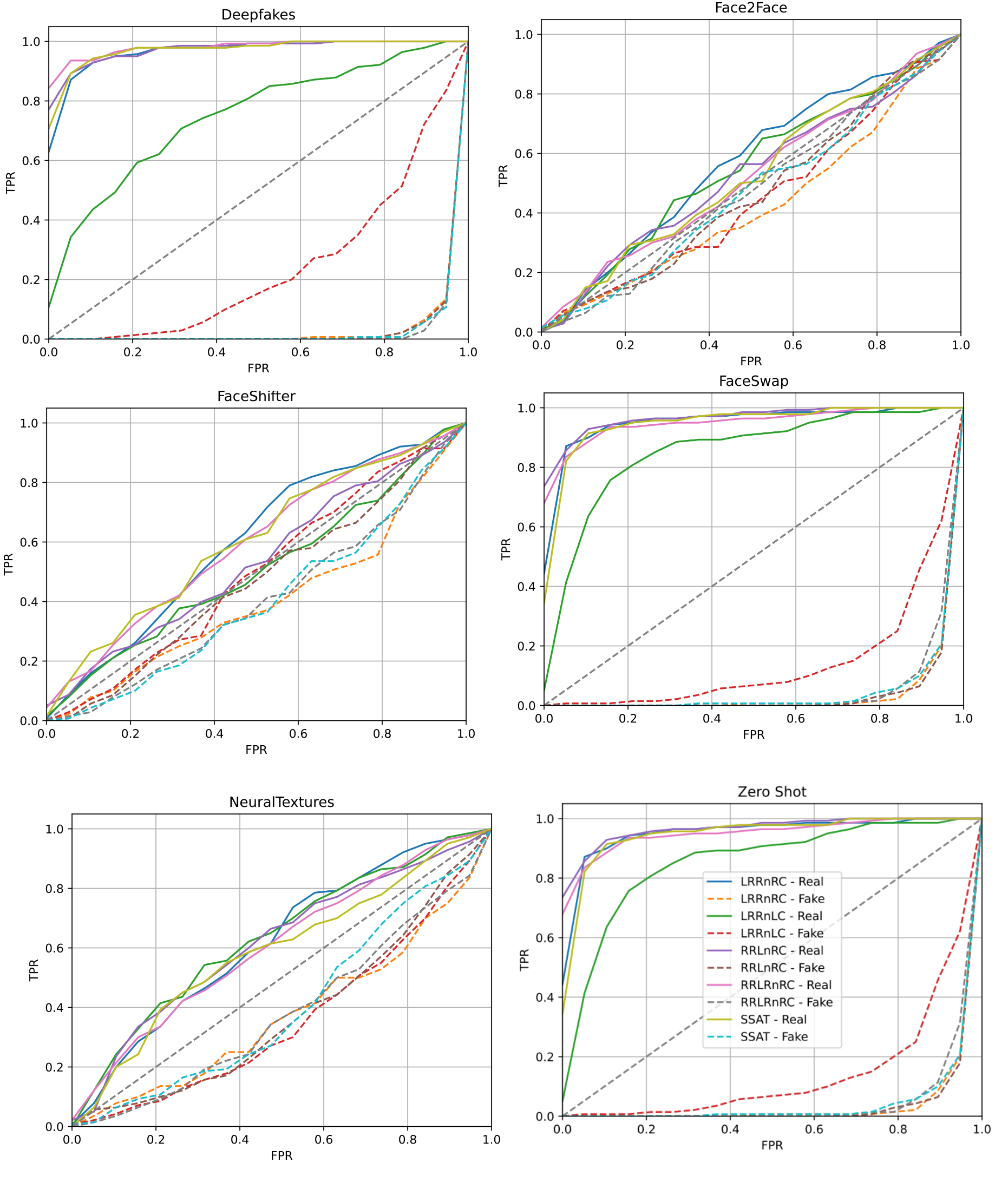}
  \vspace{-3mm}
  \caption{ROCs for different manipulation techniques and zero-shot experiments on deepfake}
\end{figure*}
\vspace{-4mm}
\subsection{Datasets}
\vspace{-2mm}
Following are the datasets used for experimentation: \noindent \textbf{FaceForensics++\cite{roessler2019faceforensicspp}} is a huge benchmark dataset to help build automated algorithms that can identify deepfakes and other face modifications on a large scale. The dataset contains over 1,000 high-quality videos generated using facial reenactment, face swapping, and deepfake generation with over 500,000 frames. The dataset's videos are divided into four groups: Deepfakes, Face2Face, FaceSwap, and NeuralTextures. 
\noindent \textbf{DFDC\cite{DFDC2020}} Deepfake Detection Challenge (DFDC) is a large dataset used to test how well fakes can be found. There are over 100,000 films in the collection that were made using different face-changing methods. There are also 140,000 movies and eight algorithms. 
\noindent\textbf{CelebA\cite{liu2015faceattributes}} is a large collection of facial features with more than 200,000 celebrity images and their corresponding 40 attribute annotations. The CelebA dataset consists of 10,177 personal identities, 202,599 facial images, 5 landmark locations, and 40 binary attribute annotations per image.
\noindent\textbf{Affectnet\cite{8013713}} utilizes a dataset of over 1,000,000 face photographs, with half of them classified for seven emotions and their corresponding valence/arousal intensity. 
\vspace{-3mm}
\subsection{Implementation details}
\vspace{-3mm}
The model is trained on images/videos of dimensions 224x224 pixels, using
patches of 16x16 pixels as input to the model. In batches of 8, the training
is performed on an NVIDIA A100 GPU using the Stochastic Gradient Descent optimizer for weight modifications. The learning rate is established by a cosine schedule that commences at 0.00005 and aims to achieve a minimum learning rate of 1e-6. The learning rate is systematically reduced by integrating a decay rate of 0.05 during each of the 75 epochs of training that the model undergoes. The stochastic depth is incorporated using an initial drop path rate of 0.01, and the various loss terms are equalized using a Lambda \(\lambda\) value of 0.1. A masking ratio of 0.75 is established. For all the datasets the official partitions are used.

\vspace{-2mm}

\subsection{Experimental Results}
\vspace{-2mm}
Tables 1, 2, 3, 4 and 5 summarize the experimental results. Table 1 and 2 highlights the model's effectiveness in detecting deepfakes, with the RRLnRC approach outperforming all others. Similarly, Table 3 shows that methods RRLRnRC consistently achieve strong performance in facial attribute categorization using CelebA, particularly for attributes like Wavy Hair and Pointy Nose. Tables 4 and 5 the approaches that incorporate local patterns show a better performance for emotion recognition as compared to the approaches solely relying on RGB. We also experimented with the methodology concerning fairness or bias in both demographic and non-demographic scenarios such as deepfake detection. We can see that our model has better generalise than SSAT. Also for demographic scenarios such as (male or not in Table 3), we achieved better results in terms of fairness (0.93 for males and 0.88 for not males whereas SSAT 0.88 for males and 0.79 not males). 
\vspace{-5mm}
\begin{table}[]
\centering
\caption{Accuracy of various methods in Deepfake classification using local pattern as LDP with SSAT trained on FaceForensics++\cite{roessler2019faceforensicspp}.}
\resizebox{\textwidth}{!}{%
\begin{tabular}{|c|c|c|c|c|c|c|}
\hline
\multicolumn{7}{|c|}{\textbf{LDP-SSAT Trained on FaceForensics++}} \\ \hline
\multirow{2}{*}{\textbf{Method}} & \multicolumn{6}{|c|}{\textbf{Accuracy}} \\ \cline{2-7} 
 & \textbf{Deepfakes} & \textbf{Face2Face} & \textbf{FaceSwap} & \textbf{NeuralTextures} & \textbf{FaceShifter} & \textbf{Avg-Accuracy} \\ \hline
\multicolumn{1}{|l|}{\begin{tabular}[c]{@{}l@{}}LRRnRC\end{tabular}} & 0.88 & 0.65 & 0.89 & 0.69 & 0.68 & 0.76 \\ \hline
\multicolumn{1}{|l|}{\begin{tabular}[c]{@{}l@{}}LRRnLC\end{tabular}} & 0.73 & 0.58 & 0.73 & 0.51 & 0.55 & 0.62 \\ \hline
\multicolumn{1}{|l|}{\begin{tabular}[c]{@{}l@{}}\textbf{RRLnRC}\end{tabular}} & 0.88 & 0.69 & 0.87 & 0.7 & 0.69 & \textbf{0.77}\\ \hline
\multicolumn{1}{|l|}{\begin{tabular}[c]{@{}l@{}}RRLRnRC\end{tabular}} & 0.88 & 0.67 & 0.85 & 0.71 & 0.69 & 0.76\\ \hline
\multicolumn{1}{|l|}{\begin{tabular}[c]{@{}l@{}}SSAT\end{tabular}} & 0.88 & 0.65 & 0.88 & 0.73 & 0.68 & 0.76 \\ \hline
\multicolumn{1}{|l|}{\begin{tabular}[c]{@{}l@{}}MAE\end{tabular}} & 0.86 & 0.66 & 0.63 & 0.63 & 0.66 & 0.68 \\ \hline
\end{tabular}%
}
\end{table}
\vspace{-9mm}

\begin{table}[]
\centering
\caption{Accuracy and AUC of various methods in Deepfake on zero-shot classification using local pattern as LDP with SSAT trained on FaceForensics++ \cite{roessler2019faceforensicspp}}
\resizebox{\textwidth}{!}{%
\begin{tabular}{|c|c|c|c|c|c|c|c|c|c|c|c|c|}
\hline
\multicolumn{13}{|c|}{\textbf{LDP-SSAT Trained on FaceForensics++ (Zero Shot)}} \\ \hline
\multirow{2}{*}{\textbf{Method}} & \multicolumn{2}{|c|}{\textbf{Deepfakes}} & \multicolumn{2}{|c|}{\textbf{Face2Face}} & \multicolumn{2}{|c|}{\textbf{FaceSwap}} & \multicolumn{2}{|c|}{\textbf{NeuralTextures}} & \multicolumn{2}{|c|}{\textbf{FaceShifter}} & \multicolumn{2}{|c|}{\textbf{Avg-Accuracy}} \\ \cline{2-13} 
 & \textbf{Acc.} & \textbf{AUC} & \textbf{Acc.} & \textbf{AUC} & \textbf{Acc.} & \textbf{AUC} & \textbf{Acc.} & \textbf{AUC} & \textbf{Acc.} & \textbf{AUC} & \textbf{Acc.} & \textbf{AUC} \\ \hline
\multicolumn{1}{|l|}{\begin{tabular}[c]{@{}l@{}}\textbf{LRRnRC}\end{tabular}} & 0.90 & 0.97 & 0.52 & 0.57 & 0.90 & 0.96 & 0.56 & 0.62 & 0.53 & 0.60 & 0.68 & \textbf{0.74} \\ \hline
\multicolumn{1}{|l|}{\begin{tabular}[c]{@{}l@{}}LRRnLC\end{tabular}} & 0.69 & 0.76 & 0.54 & 0.55 & 0.80 & 0.86 & 0.60 & 0.64 & 0.53 & 0.51 & 0.63 & 0.66 \\ \hline
\multicolumn{1}{|l|}{\begin{tabular}[c]{@{}l@{}}RRLnRC\end{tabular}} & 0.88 & 0.98 & 0.55 & 0.53 & 0.89 & 0.97 & 0.60 & 0.62 & 0.54 & 0.54 & 0.69 & 0.73 \\ \hline
\multicolumn{1}{|l|}{\begin{tabular}[c]{@{}l@{}}RRLRnRC\end{tabular}} & 0.90 & 0.98 & 0.54 & 0.53 & 0.88 & 0.95 & 0.57 & 0.61 & 0.55 & 0.60 & 0.69 & 0.73 \\ \hline
\multicolumn{1}{|l|}{\begin{tabular}[c]{@{}l@{}}\textbf{SSAT}\end{tabular}} & 0.88 & 0.98 & 0.53 & 0.53 & 0.86 & 0.96 & 0.59 & 0.60 & 0.57 & 0.61 & 0.69 & \textbf{0.74} \\ \hline
\end{tabular}%
}
\end{table}

\vspace{-4mm}

\begin{table}[]
\centering
\caption{Accuracy of various methods in emotion classification using local pattern as LDP with SSAT trained on CelebA\cite{liu2015faceattributes}.}
\resizebox{\textwidth}{!}{%
\begin{tabular}{|lccccccc|}
\hline
\multicolumn{8}{|c|}{\textbf{LDP-SSAT Trained on CelebA}} \\ \hline
\multicolumn{1}{|c|}{\multirow{2}{*}{\textbf{Method}}} & \multicolumn{7}{c|}{\textbf{Accuracy}} \\ \cline{2-8} 
\multicolumn{1}{|c|}{} & \multicolumn{1}{c|}{\textbf{Oval Face}} & \multicolumn{1}{c|}{\textbf{Wavy Hair}} & \multicolumn{1}{c|}{\textbf{Pointy Nose}} & \multicolumn{1}{c|}{\textbf{Mustache}} & \multicolumn{1}{c|}{\textbf{Eye Glasses}} & \multicolumn{1}{c|}{\textbf{Male}} & \multicolumn{1}{l|}{\textbf{Avg-Accuracy}} \\ \hline
\multicolumn{1}{|l|}{LRRnRC} & \multicolumn{1}{c|}{0.74} & \multicolumn{1}{c|}{0.83} & \multicolumn{1}{c|}{0.77} & \multicolumn{1}{c|}{0.96} & \multicolumn{1}{c|}{0.91} & \multicolumn{1}{c|}{0.89} & 0.85 \\ \hline
\multicolumn{1}{|l|}{LRRnLC} & \multicolumn{1}{c|}{0.73} & \multicolumn{1}{c|}{0.80} & \multicolumn{1}{c|}{0.74} & \multicolumn{1}{c|}{0.96} & \multicolumn{1}{c|}{0.99} & \multicolumn{1}{c|}{0.86} & 0.84 \\ \hline
\multicolumn{1}{|l|}{\textbf{RRLnRC}} & \multicolumn{1}{c|}{0.74} & \multicolumn{1}{c|}{0.82} & \multicolumn{1}{c|}{0.76} & \multicolumn{1}{c|}{0.96} & \multicolumn{1}{c|}{0.99} & \multicolumn{1}{c|}{0.94} & \textbf{0.87} \\ \hline
\multicolumn{1}{|l|}{\textbf{RRLRnRC}} & \multicolumn{1}{c|}{0.74} & \multicolumn{1}{c|}{0.84} & \multicolumn{1}{c|}{0.77} & \multicolumn{1}{c|}{0.96} & \multicolumn{1}{c|}{0.99} & \multicolumn{1}{c|}{0.92} &\textbf{ 0.87} \\ \hline
\multicolumn{1}{|l|}{SSAT} & \multicolumn{1}{c|}{0.75} & \multicolumn{1}{c|}{0.82} & \multicolumn{1}{c|}{0.76} & \multicolumn{1}{c|}{0.96} & \multicolumn{1}{c|}{0.99} & \multicolumn{1}{c|}{0.81} & 0.84 \\ \hline
\multicolumn{1}{|l|}{MAE} & \multicolumn{1}{c|}{0.67} & \multicolumn{1}{c|}{0.63} & \multicolumn{1}{c|}{0.62} & \multicolumn{1}{c|}{0.74} & \multicolumn{1}{c|}{0.78} & \multicolumn{1}{c|}{0.62} & 0.67 \\ \hline
\end{tabular}%
}
\end{table}

\vspace{-5mm}
\begin{table}[h]
\centering
\caption{Accuracy of various methods in emotion classification using local pattern as LDP with SSAT trained on Affectnet\cite{8013713}.}
\resizebox{\textwidth}{!}{%
\begin{tabular}{|c|c|c|c|c|c|c|c|c|c|c|}
\hline
\multicolumn{11}{|c|}{\textbf{LDP-SSAT Trained on Affectnet}} \\ \hline
\multirow{2}{*}{\textbf{Method}} & \multicolumn{10}{|c|}{\textbf{Accuracy}} \\ \cline{2-11} 
 & \textbf{Neutral} & \textbf{Happy} & \textbf{Sad} & \textbf{Surprise} & \textbf{Fear} & \textbf{Disgust} & \textbf{Anger} & \textbf{Contempt} & \textbf{Non-Face} & \textbf{Avg-Accuracy} \\ \hline
\multicolumn{1}{|l|}{\begin{tabular}[c]{@{}l@{}}\textbf{LRRnRC}\end{tabular}} & 0.74 & 0.53 & 0.91 & 0.95 & 0.97 & 0.98 & 0.91 & 0.98 & 0.99 & \textbf{0.88} \\ \hline
\multicolumn{1}{|l|}{\begin{tabular}[c]{@{}l@{}}\textbf{LRRnLC}\end{tabular}} & 0.74 & 0.53 & 0.91 & 0.95 & 0.97 & 0.98 & 0.91 & 0.98 & 0.99 & \textbf{0.88} \\ \hline
\multicolumn{1}{|l|}{\begin{tabular}[c]{@{}l@{}}\textbf{RRLnRC}\end{tabular}} & 0.74 & 0.52 & 0.90 & 0.94 & 0.97 & 0.98 & 0.90 & 0.98 & 0.99 & \textbf{0.88} \\ \hline
\multicolumn{1}{|l|}{\begin{tabular}[c]{@{}l@{}}\textbf{RRLRnRC}\end{tabular}} & 0.74 & 0.53 & 0.91 & 0.95 & 0.97 & 0.98 & 0.91 & 0.98 & 0.99 & \textbf{0.88} \\ \hline
\multicolumn{1}{|l|}{\begin{tabular}[c]{@{}l@{}}SSAT\end{tabular}} & 0.74 & 0.52 & 0.90 & 0.94 & 0.96 & 0.97 & 0.90 & 0.97 & 0.98 & 0.87\\ \hline
\multicolumn{1}{|l|}{\begin{tabular}[c]{@{}l@{}}MAE\end{tabular}} & 0.72 & 0.51 & 0.90 & 0.95 & 0.95 & 0.98 & 0.91 & 0.98 & 0.97 & 0.87\\ \hline
\end{tabular}%
}
\end{table}
\vspace{10mm}
\begin{table}[ht!]
\centering
\caption{Accuracy of various methods in emotion classification using local pattern as LBP with SSAT trained on Affectnet\cite{8013713}.}
\resizebox{\textwidth}{!}{%
\begin{tabular}{|c|c|c|c|c|c|c|c|c|c|c|}
\hline
\multicolumn{11}{|c|}{\textbf{LBP-SSAT Trained on Affectnet}} \\ \hline
\multirow{2}{*}{\textbf{Method}} & \multicolumn{10}{|c|}{\textbf{Accuracy}} \\ \cline{2-11} 
 & \textbf{Neutral} & \textbf{Happy} & \textbf{Sad} & \textbf{Surprise} & \textbf{Fear} & \textbf{Disgust} & \textbf{Anger} & \textbf{Contempt} & \textbf{Non-Face} & \textbf{Avg-Accuracy} \\ \hline
\multicolumn{1}{|l|}{\begin{tabular}[c]{@{}l@{}}\textbf{LRRnRC}\end{tabular}} & 0.74 & 0.52 & 0.91 & 0.95 & 0.97 & 0.98 & 0.9 & 0.98 & 0.99 & \textbf{0.88} \\ \hline
\multicolumn{1}{|l|}{\begin{tabular}[c]{@{}l@{}}\textbf{LRRnLC}\end{tabular}} & 0.74 & 0.53 & 0.91 & 0.95 & 0.97 & 0.98 & 0.91 & 0.98 & 0.99 & \textbf{0.88} \\ \hline
\multicolumn{1}{|l|}{\begin{tabular}[c]{@{}l@{}}\textbf{RRLnRC}\end{tabular}} & 0.74 & 0.52 & 0.90 & 0.94 & 0.97 & 0.98 & 0.90 & 0.98 & 0.99 & \textbf{0.88} \\ \hline
\multicolumn{1}{|l|}{\begin{tabular}[c]{@{}l@{}}RRLRnRC\end{tabular}} & 0.75 & 0.51 & 0.89 & 0.93 & 0.95 & 0.96 & 0.89 & 0.96 & 0.97 & 0.87 \\ \hline
\multicolumn{1}{|l|}{\begin{tabular}[c]{@{}l@{}}SSAT\end{tabular}} & 0.74 & 0.52 & 0.90 & 0.94 & 0.96 & 0.97 & 0.90 & 0.97 & 0.98 & 0.87\\ \hline
\multicolumn{1}{|l|}{\begin{tabular}[c]{@{}l@{}}MAE\end{tabular}} & 0.72 & 0.5 & 0.91 & 0.92 & 0.97 & 0.95 & 0.91 & 0.94 & 0.99 & 0.86\\ \hline
\end{tabular}%
}
\end{table}
\vspace{-8mm}

ROCs for different manipulation techniques and zero-shot experiments on deepfake are in Fig 3. The experimental results demonstrate that the proposed methods yield substantial enhancements in the performance of numerous tasks. Compared to conventional RGB reconstruction, local pattern reconstruction offers several advantages. It facilitates the acquisition of intricate structure data and localized texture heterogeneity, which is crucial for applications such as emotion categorization, face feature identification, and deepfake identification.

In contrast to RGB reconstruction, which emphasizes pixel-level intensity, local pattern methods prioritize structural and textural patterns and force the mask auto-encoder to learn better structural patterns. As a result achieved better performance on face analysis on different tasks. These qualities increase their trustworthiness and appropriateness in numerous real-world situations by making them more resistant to changes in lighting, obstructions, and slight changes in facial expression.
 
\vspace{-4mm}
\section{Conclusion}
\vspace{-3mm}
This work introduces a hybrid method that integrates texture-based and model-based features with self-supervised auxiliary tasks to improve the efficiency of facial analytic models in a Vision Transformer design. Using local pattern elements in SSAT provides notable benefits compared to conventional RGB-based approaches. This is due to its ability to accurately reconstruct complex structural and textural patterns, therefore improving the model's ability to withstand facial patterns for different tasks. 
\vspace{-2mm}
\section*{Acknowledgements}
\vspace{-2mm}
This work was funded by the Institute of Data Engineering, Analytics, and Science (IDEAS) Technology Innovation Hub (TiH) Indian Statistical Institute, Kolkata, Department of Science and Technology (DST), Government of India under the project titled "Generalized Tampering Detection in Media (GTDM)" and project number OO/ISI/IDEAS-TIH/2023-24/86.
\vspace{-4mm}
\bibliographystyle{plain}
\bibliography{acmart}
\end{document}